\documentclass[runningheads]{llncs}
\usepackage[T1]{fontenc}
\usepackage{graphicx}
\usepackage{booktabs}
\usepackage[misc]{ifsym}

\usepackage{amsmath,mathtools, amssymb}
\usepackage{algorithm, algorithmic}
\usepackage{xcolor}
\usepackage{caption}
\usepackage{subcaption}
\usepackage{cite}
\usepackage{dsfont}   % to use indicator function
\usepackage{tablefootnote}  % Footnote in tables

\definecolor{internalnodecolor}{HTML}{337DCC}   % Lighter Blue
\definecolor{leafnodecolor}{HTML}{FFA500}   % Orange
\usepackage{tikz}
\usetikzlibrary{positioning}
\usepackage{pgfplots}
\usepackage{pgfplotstable}
\pgfplotsset{compat=newest}
\tikzset{
  rectanglenode/.style={
    draw
  , align=center
  , execute at begin node=\setlength{\baselineskip}{0.8em}
  }
}
\pgfdeclarelayer{foreground} 
\pgfdeclarelayer{background}
   \pgfsetlayers{background,%
                 main,%
                 foreground%
                 }

\begin{document}

\title{Soft Hoeffding Tree: A Transparent and Differentiable Model on Data Streams}

\titlerunning{Soft Hoeffding Tree}
% If the full title of your paper is short enough to also fit in the running head, you can omit the abbreviated paper title here. You can check as follows: if you comment out the \titlerunning line, something will appear in the header of all odd-numbered pages of your PDF from page 3 onward. This something is either the full title (in which case all is well), or the error message "Title Suppressed Due to Excessive Length". If this error message appears, you're going to want to provide an abbreviated title within the \titlerunning command, because if you won't do it, Springer will do it for you.

\author{Kirsten K{\"o}bschall\orcidID{0009-0001-3967-0891} \and
Lisa Hartung\orcidID{0000-0002-3420-551X} \and
Stefan Kramer\orcidID{0000-0003-0136-2540}
}
% You may leave out the orcidID information, if you want to.
% Use \corr to indicate the corresponding author. Note the spacing around the \corr command. Only one author can be the corresponding author.

\authorrunning{K. K{\"o}bschall et al.}
% First names are abbreviated in the running head.
% If there is one author, write 'A.L. Benjamin'.
% If there are two authors, write 'A.L. Benjamin and C.C. Broadus Jr.'
% If there are more than two authors, '[...] et al.' is used.

\institute{Johannes Gutenberg University Mainz, Mainz, Germany \\
\email{koebschall@uni-mainz.de}
}

\maketitle              % typeset the header of the contribution

\begin{abstract}
% 150--250 words.
We propose soft Hoeffding trees (SoHoT) as a new differentiable and transparent model for possibly infinite and changing data streams. Stream mining algorithms such as Hoeffding trees grow based on the incoming data stream, but they currently lack the adaptability of end-to-end deep learning systems. End-to-end learning can be desirable if a feature representation is learned by a neural network and used in a tree, or if the outputs of trees are further processed in a deep learning model or workflow. Different from Hoeffding trees, soft trees can be integrated into such systems due to their differentiability, but are neither transparent nor explainable. Our novel model combines the extensibility and transparency of Hoeffding trees with the differentiability of soft trees. We introduce a new gating function to regulate the balance between univariate and multivariate splits in the tree. Experiments are performed on 20 data streams, comparing SoHoT to standard Hoeffding trees, Hoeffding trees with limited complexity, and soft trees applying a sparse activation function for sample routing. The results show that soft Hoeffding trees outperform Hoeffding trees in estimating class probabilities and, at the same time, maintain transparency compared to soft trees, with relatively small losses in terms of AUROC and cross-entropy. We also demonstrate how to trade off transparency against performance using a hyperparameter, obtaining univariate splits at one end of the spectrum and multivariate splits at the other.

\keywords{Decision tree \and Data streams \and Hoeffding bound \and Soft trees \and Concept drift \and Differentiability.}
%\keywords{Stream learning \and Decision tree \and Hoeffding bound \and Soft trees \and Differentiability.}
\end{abstract}

\section{Introduction}
% Background and motivation
More than twenty years after the introduction of the first stream mining algorithms, the data stream setting and variants are still gaining momentum and importance, considering the recent interest in real-time AI.
% As the amount of data continues to grow, data stream environments are becoming increasingly common in various industries. Especially since resources are limited nowadays and data should be processed but cannot be stored.
Moreover, since AI models have become indispensable in everyday life, it has become more relevant to build trust and ensure ethical and responsible behavior in machine learning supported decision processes. One of the desirable properties of machine learning in general, and thus also of machine learning on data streams, is transparency, i.e. that a model is in principle interpretable by human users.

% Even transparency can help identify false or unfair decisions.
% Related work, Problem description
% Hoeffding Tree -> Soft Tree -> Soft Hoeffding Tree
Hoeffding trees \cite{Domingos2000} are among the most popular choices when working with data streams. The possibility to inspect the structure and importance of variables of decision trees is frequently considered a benefit of the approach. However, trees lack a good mechanism for representation learning, at which neural networks excel \cite{Bengio2013}. Due to discontinuities in the loss function of decision trees caused by their discrete splits, standard Hoeffding trees cannot be integrated in a neural network.
Soft trees are a differentiable version of decision trees which allow optimization using gradient-based methods. Hazimeh et al. \cite{hazimeh2020tree} introduced conditional computation on soft trees and propose an ensemble of soft trees as a new layer for neural networks. 

\begin{figure}[t!]
\centering
% To scale add scale=0.8 
\scalebox{0.65}{
    \includegraphics[]{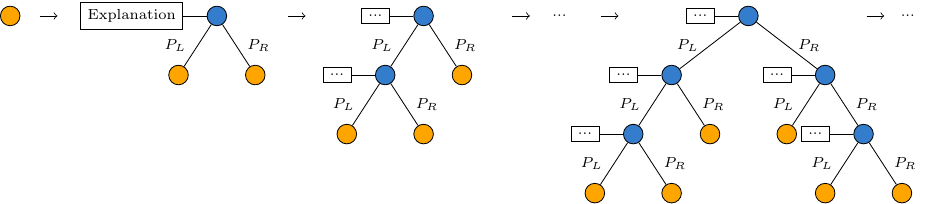}
}
\caption{SoHoT adaptation to an evolving data stream over time, denoted by routing probabilities to the left child ($P_L$) or right child ($P_R$).}
\label{fig:growth_sohot}
\end{figure}
To leverage the strengths of both concepts, we introduce the  soft Hoeffding tree (SoHoT), a new differentiable and transparent model on streaming data.
%As we do not have information of the data in streaming environments, it is difficult to define a tree structure in advance like it is done for standard soft trees.
In order to enable our model to dynamically adjust the tree structure, to accommodate to evolving data streams, and enable a transparent prediction, we combine soft trees with the concept of Hoeffding trees as visualized in Figure~\ref{fig:growth_sohot}.

% Summarise our contributions: SoHoTs with transparent rouing function and growing a tree structure based on streams
The main contributions of this paper are: (i) We propose a new transparent routing function and (ii) show how to split nodes in a gradient-based tree in a data stream setting by applying the Hoeffding inequality, and (iii) introduce a metric to measure feature importance in a soft tree and soft Hoeffding tree.

% TODO: Evaluation and Results
The results of SoHoTs are presented in comparison to Hoeffding trees with an unlimited and a limited number of decision nodes, and to soft trees: First, SoHoTs outperform Hoeffding trees in estimating class probabilities. Second, we shed light on the trade-off between transparency and performance in a SoHoT through a hyperparameter of the proposed gating function. Soft trees have a slightly better performance compared to SoHoTs, but our analysis reveals that SoHoTs are more transparent, as evidenced by a metric assessing feature importance for decision rules in the tree. 

% Paper structure
This paper is structured as follows: Related work is discussed in Section~\ref{sec:relatedwork}.
Section~\ref{sec:sohot} presents the new method, Soft Hoeffding Trees (SoHoT).
An explanation of the transparency property is given in Section~\ref{sec:transparency}. The results are presented in Section~\ref{sec:experiments}, and finally, Section~\ref{sec:conclusion} concludes our work with an outlook on further research.

% Section: Related Work
\section{Related Work}\label{sec:relatedwork}
% Story: Streamable tree (Hoeffding trees) but no integration in NN
%       Soft trees are differentiable
%       But not transparent

% ------------- Variants of Hoeffding Trees -------------
Hoeffding trees (HT) \cite{Domingos2000}, as introduced by Domingos and Hulten in 2000, are an algorithm for mining decision trees from continuously changing data streams, which is still the basis for many state-of-the-art learners for data streams \cite{Gouk2019, SurveySCL2023}. We briefly review the principle of HT in order to apply some individual techniques in Section~\ref{sec:sohot}.
In contrast to standard decision trees, HT uses the Hoeffding bound to determine whether there is sufficient evidence to select the best split test, or if more samples are needed to extend the tree.
More precisely, let $G(x_i)$ be the heuristic measure (e.g. information gain) for an attribute $x_i$ to evaluate for a split test, and $\bar{G}(x_i)$ the observed value after $n$ samples. A Hoeffding tree guarantees with probability $1-\delta$ that the attribute chosen after seeing $n$ samples is the same as if an infinite number of samples had been seen. 
If $\bar{G}(x_a)-\bar{G}(x_b)>\epsilon$, where $x_a$ is the attribute with the highest observed $\bar{G}$ after $n$ samples, $x_b$ the second-best attribute, then $x_a$ is the best attribute to perform a split on a leaf node with probability $1-\delta$, where $\delta$ is the significance level, $\epsilon = \sqrt{(R^2\ln{1/\delta})/(2n)}$ and $R=\log k$ and $k$ the number of classes.
Hulten et al. proposed the Concept-adapting Very Fast Decision Tree learner (CVFDT) \cite{Hulten2001} to adapt to changing data streams by building an alternative subtree and replacing the old with the new as soon as the old subtree becomes less accurate.
% Adaptive Learning from Evolving Data Streams - Bifet, Gavalda
Gavald\`{a} and Bifet \cite{GavaldBifet2009} proposed the Hoeffding window tree and the Hoeffding adaptive tree as a sliding window approach and an adaptive approach, respectively, to deal with distribution and concept drift based on change detectors and estimator modules. 
The Hoeffding window tree maintains a sliding window of instances, and the Hoeffding adaptive tree overcomes the issue of having to choose a window size by storing instances of estimators of frequency statistics at each node.
Nevertheless, decision trees (esp. Hoeffding trees) are non-differentiable, since hard routing (i.e., a sample can only be routed as a whole to the left or the right) causes discontinuities in the loss function \cite{hazimeh2020tree},  making them incompatible with end-to-end learning and therefore unsuitable for integration into neural networks. In deep learning pipelines, where a feature representation is learned, it is desirable to use differentiable models to enable end-to-end learning.
%In order to adapt the existing tree to a new concept, this can only be done by keep growing or replacing (sub-)trees.
% ------------- Variants of Perceptron Trees -------------
%Perceptron decision trees were introduced by Utgoff \cite{1988Utgoff} as a decision tree where each leaf node is a linear threshold unit, i.e., a perceptron. Zhou \cite{2002Zhou} combined decision trees and neural networks by presenting hybrid decision trees (HDL). Each leaf node in a HDL embeds a feedforward neural network.
% Naive Bayes: Stress-testing HT
Naive Bayes Hoeffding trees are a hybrid adaptive method having trees with naive Bayes models in the leaf nodes \cite{2005HolmesNaiveBayes}.
For each training sample, a naive Bayes prediction is made and compared to the majority class voting. 
% Perceptron HT
Another variant of Hoeffding trees is to replace the model in the leaf nodes by a perceptron classifier \cite{2010BifetPerceptron}. %The latter approach minimizes runtime  without compromising its highly competitive accuracy.
% ------------- Ensemble Methods -------------
Ensemble methods for data streams \cite{onlineensemble2001} are widely-used and include algorithms like adaptive random forests \cite{ARF2017}, which feature effective resampling and adaptive operators such as drift detection and recovery strategies.
In contrast to this line of work, we focus on individual trees in this paper. The study of the behavior in ensembles will be investigated in future work.
%as a strong base learner is a requirement for the effectiveness of any ensemble method. CITE?

% ------------- Variants of Soft Decision Trees -------------
Since hard-routing decision trees lack a good mechanism for representation learning \cite{hazimeh2020tree}, soft trees (ST) were introduced as differentiable decision trees. Initially proposed by Jordan and Jacobs \cite{JordanJacobs1994}, they were further developed in various directions by other researchers \cite{Kontschieder2015,frosst2017distilling,2019Hehn}.
We briefly discuss soft trees, to be in a better position to explain individual mechanisms in the following section.
Soft trees perform soft routing, i.e. an internal node distributes a sample simultaneously to both the left and the right child, allowing for different proportions in each direction. A common choice for the gating function is the sigmoid function. Hazimeh et al. \cite{hazimeh2020tree} introduced the smooth-step function $S$ as a continuously differentiable gating function, i.e.,
\begin{align}
	S(t) &=
	\begin{dcases}
		0 & t \leq -\gamma /2 \\
		-\frac{2}{\gamma^3}t^3 + \frac{3}{2\gamma}t + \frac{1}{2} & -\gamma /2 \leq t \leq \gamma /2 \\
		1 & t \geq \gamma /2,
	\end{dcases}
	\label{smoothstepfunc}
\end{align}
where $\gamma$ is a non-negative scalar. 
A property of $S$ is the ability to output exact zeros and ones, which provides a balance between soft and hard routed samples and enables a conditional computation.
Along with the smooth-step function, the authors proposed the tree ensemble layer (TEL) for neural networks as an additive model of soft trees utilizing $S$ and concluded that TEL has a ten-fold speed-up compared to differentiable trees and a twenty-fold reduction in the number of parameters compared to gradient boosted trees.
{\.I}rsoy et al. proposed an incremental architecture with soft decision trees \cite{2012Irsoy}, where the tree grows incrementally as long as it improves.
Hehn et al. \cite{2019Hehn} proposed a greedy algorithm to grow a tree level-by-level by splitting nodes, optimizing it by maximizing the log-likelihood on subsets of the training data.
%Each split in the tree is a Bernoulli decision and the maximization objective of the training data is the empirical log-likelihood.
%Differential decision trees does not have initially the ability to learn from data streams and dynamically build the tree structure based on the incoming data stream.
% ------------- Stochastic gradient trees - Gouk, Pfahringer and Frank -------------
Stochastic gradient trees \cite{Gouk2019} are an incremental learning algorithm using stochastic gradient information to evaluate splits and compute leaf node predictions. The tree applies $t$-tests instead of the Hoeffding inequality to decide whether a node is to be split. 
%Enabling classification, regression or multi-instance learning, only requires a change of the loss function.
Generally, differentiable decision trees commonly lack transparency and the facility to adjust to concept drift. Finally, they do not have the ability to learn and dynamically build the model's architecture from a data stream.

% Section: Soft Hoeffding Tree
\section{Soft Hoeffding Tree}\label{sec:sohot}
%A Soft Hoeffding Tree (SoHoT) is a transparent and differentiable decision tree for extremely large datasets, especially for streaming data environments. Our proposed model takes advantage of both the extensibility and interpretability of Hoeffding trees, and the differentiability and the new routing mechanism using the smooth-step function $S$ of soft trees.
A soft Hoeffding tree is a transparent and differentiable decision tree for data streams.
The tree is tested and trained per instance or mini-batch and can generate predictions at any point in time during the sequential processing.
% Since the data stream is handled in a sequential manner, there is no need to store the data, resulting in substantial resource savings.
We consider a supervised learning setting, an input space $\mathcal{X} \subseteq \mathbb{R}^p$ and an output space $\mathcal{Y} \subseteq \mathbb{R}^k$, where $k$ equals the number of classes.

\subsection{Definition}
A function $T:\mathcal{X}\rightarrow \mathbb{R}^k$ is called a soft Hoeffding tree, if $T$ fulfils the properties of an HT with a new routing function (Eq.~\ref{eq:routing}). Consequently, the training of $T$ differs from the training of an HT, which we will discuss in Section~\ref{subsec:adaption}.
Let $x \in \mathbb{R}^p$ be an input sample, and $\mathcal{I}$ and $\mathcal{L}$ denote a set of internal nodes and leaf nodes, respectively.
Each internal node $i\in \mathcal{I}$ has a weight $w\in \mathbb{R}^p$ and holds a split decision, and each leaf node $l\in \mathcal{L}$ has a weight $o\in \mathbb{R}^k$ and holds sufficient statistics to compute the heuristic function $G$ as explained in Section~\ref{sec:relatedwork}.
Moreover, let $\mathcal{W} \coloneqq \{w_i \mid i \in \mathcal{I}\}$ and $\mathcal{O} \coloneqq \{o_l \mid l \in \mathcal{L}\}$.
%Analogously to TEL \cite{hazimeh2020tree}, we allow an ensemble of $m$ soft Hoeffding trees and use it as a standalone layer. Let $T^{(j)}:\mathcal{X}\rightarrow \mathbb{R}^k$ be the $j$th tree of a tree ensemble of $m$ trees.
%The output of the layer is computed as follows: $\mathcal{T}(x) = T^{(1)}(x) + T^{(2)}(x) + \dots + T^{(m)}(x)$.
For classification tasks, the softmax function can be applied to the output of $T$ to obtain class probabilities.
%For classification tasks, applying a softmax to the raw predictions yields class probabilities.
%Figure~\ref{fig:ensemble_sht} visualizes an ensemble of SoHoTs. 
%The node weights are initialized randomly at the beginning, and after each new split (see Section~\ref{subsec:adaption}), therefore the routing probabilities for a fixed instance $x$ can differ between each initialized tree.
We begin by introducing the forward pass to determine a prediction for an input, followed by the backward pass to show how the tree adapts to drifts.
At the beginning, $T$ consists of a single leaf node.

\subsection{Prediction}\label{ssht:prediction}
% Argumentation structure:
%   1. New gating function
%   2. Explain forward pass
The input $x$, starting at the root node $i$, is routed to the left child (if it exists), which is denoted by $\{x \swarrow i\}$, with probability $ P(\{x \swarrow i\})$ and to the right child (if it exists) with probability $ P(\{x \searrow i\}) = 1- P(\{x \swarrow i\})$. 
% Introduce new gating function
To achieve both a transparent model and a transparent prediction, a new routing mechanism is proposed.
We introduce a convex combination of the split test computed for an internal node and the routing probability by applying the smooth-step function $S$ (Eq.~\ref{smoothstepfunc}).
Let $\alpha \in [0,1]$ be a parameter, $x_a$ the split feature, $\theta$ the split value and ``$(x_a < \theta)$?'' the split test computed for the internal node $i$. Further $\mathds{1}(\cdot)$ be the indicator function, which is 1 if the condition in $(\cdot)$ is true and 0 otherwise.
Then our new routing function is defined as follows:
\begin{align}\label{eq:routing}
    P(\{x \swarrow i\}) = \alpha \cdot S(\langle w_i, x\rangle ) + (1-\alpha) \cdot \mathds{1}(x_a < \theta), 
\end{align}
where $w_i \in \mathbb{R}^p$ is the weight vector of $i$.
\begin{figure}[t!]
\centering
% To scale add scale=0.8 
\scalebox{0.85}{
    \includegraphics[]{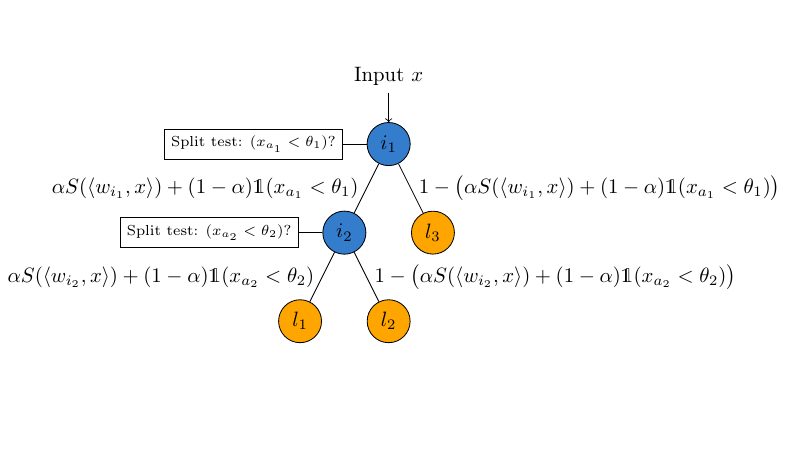}
}
\caption{Soft Hoeffding tree $T$ at time $t>0$ with routing probabilities at the edges.}
\label{fig:single_sht}
\end{figure}
In other words, the probability of routing $x$ to the left child is determined by the combined and weighted impact of the multivariate split by $S(\langle w_i, x\rangle )$ and the univariate split test by $\mathds{1}(x_a < \theta)$.
%Smaller values for $\alpha$ allow a higher transparency to the model. 
The parameter $\alpha$ regulates the transparency of the model. Specifically, when $\alpha$ becomes smaller, the model's transparency increases, as the split tests provide users with explicit information about the criterion used for decision-making. However, performing more multivariate splits may also lead to higher performance, as we evaluate in Section~\ref{sec:transparency}.
Note that with $\alpha=0$, a SoHoT acts similarly to a Hoeffding tree, and with $\alpha = 1$, a forward pass similar to soft trees is performed. More precisely, with $\alpha = 1$, the performance should resemble the performance of ST, once the tree is fully grown. More discussion of this can be found in Section~\ref{sec:transparency} and towards the end of Section~\ref{subsec:evaluation}.

We denote $P(\{x\rightarrow l\})$ as the probability that $x$ reaches the leaf $l \in \mathcal{L}$.
The final prediction for $x$ is then defined as
\begin{align}\label{formular:output}
T(x) = \sum_{l\in \mathcal{L}} P(\{x\rightarrow l\})o_l,
\end{align}
where $o_l\in \mathbb{R}^k$ is the weight vector of a leaf $l\in \mathcal{L}$.
Figure~\ref{fig:single_sht} shows an example of a SoHoT $T$ at a time step $t>0$ including the routing probabilities on each edge.
%Since the tree starts with a single leaf node it returns a result only based on the root's weight until the first split.
Note that the prediction at the beginning is only based on the root's weight due to the fact that the root $r$ is a leaf node before the first split and $P(\{x\rightarrow r\})=1$.
% Bute the weights are updated to optimize the output as discussed in~\ref{subsec:adaption}.
We also exploit the properties of $S$ and compute each root-to-leaf probability only for reachable leaves, which means $P(\{x\rightarrow i\}) > 0$ for any node $i$ on the path from the root to a leaf.
The conditional computation implies that eventually subtrees might not be visited during forward and backward pass. This makes the online framework particularly resource-efficient.
The parameter $\gamma$ in $S$ regulates the extent of conditional computation. More precisely, as the value of $\gamma$ increases, the routing becomes softer. Conversely, as $\gamma$ gets closer to $0$, the routing tends to become harder.

% Remark: Pseudo code of the forward pass is omitted here since it does not differ from TEL's forward pass (only the backward is interesting)

\subsection{Adaptation to Drifting Data}\label{subsec:adaption}
\begin{algorithm}[t]
   \caption{Backward pass}
   \label{alg:backward}
\begin{algorithmic}
    \STATE {\bfseries Input:} $T$, $x\in \mathbb{R}^p$, $\frac{\partial L}{\partial T}$, $\epsilon_s$
    \STATE {\bfseries Output:}  $\frac{\partial L}{\partial x}, \frac{\partial L}{\partial W}, \frac{\partial L}{\partial O}$
    \STATE Initialize $\frac{\partial L}{\partial x} = 0$
    \STATE Traverse $T$ in post-order:
    \begin{ALC@g}
        \STATE $i$ is the current node
        \IF{$i$ is a leaf}
            \STATE Compute $\frac{\partial L}{\partial o_i}$
            \IF{{$i$.depth $\leq$ max\_depth} \AND {$P(\{x\rightarrow i\}) > \epsilon_s$}}
                \STATE Update leaf node statistics
            \ENDIF
        \ELSE
            \STATE Compute $\frac{\partial L}{\partial x_i}$ and $\frac{\partial L}{\partial w_i}$
        \ENDIF
    \end{ALC@g}
    \STATE Iterate over all leaves and attempt to split
    \STATE {\bfseries Return:} {$\frac{\partial L}{\partial x}, \frac{\partial L}{\partial W}, \frac{\partial L}{\partial O}$}
\end{algorithmic}
\end{algorithm}
Let $L$ be a loss function.
Algorithm~\ref{alg:backward} summarizes the backward pass, wherein the model adapts to the data stream whenever a label is provided. The adaptation to the data stream can be captured in two steps: the gradient computation and the growth of the tree structure.
Given $T$, the input $x\in \mathbb{R}^p$ and $\frac{\partial L}{\partial T}$, which should be available from the backpropagation algorithm as an input to the backward algorithm.
$T$ is traversed in post-order to compute the gradients and to update the statistics in the leaf nodes.
The gradients of each weight in $\mathcal{W}$ and $\mathcal{O}$ and the input $x$ are calculated as proposed for TEL \cite{hazimeh2020tree}. 
% Update leaf node statistics
Beyond the gradient computation of the weights, we also focus on the selection of split tests within the tree. In order to determine univariate splits on a data stream, we use the Hoeffding tree methodology here. First, we collect the statistics of the incoming instances in the leaf node $l$ to compute potential split tests later, but only if the maximum depth on this path has not yet been reached and $P(\{x\rightarrow l\}) > \epsilon_s$, where $0 < \epsilon_s \leq 1$\footnote{A reasonable choice is, e.g.,  $\epsilon_s=0.25$.}. 
A cutoff $\epsilon_s$ is chosen, since certain examples have a limited likelihood of reaching leaf nodes, resulting in their inclusion in predictions with a relatively small proportion, therefore, these statistics are not relevant for any new univariate split in $l$.
% Update tree structure (attempt to split)
As soon as the tree traversal is finished, the algorithm iterates over all leaves and attempts to split (see Algorithm~\ref{alg:attempt_to_split}). 
%The statistics are temporarily stored in $i$ to compute $\bar{G}(X_j)$ for each attribute $X_j$ and to decide when to split using the Hoeffding inequality.
Determining when to split follows the same pattern as with Hoeffding trees by applying the Hoeffding inequality as explained in Section~\ref{sec:relatedwork}. The decision whether a leaf node becomes an internal node is made by applying a heuristic function $G$ (e.g., information gain) to evaluate and determine the split attribute $x_a$ and an associated split test of the form "$(x_a < \theta)$?".
In case of an extension, a leaf node $l$ evolves into a new internal node $i$ with two new child nodes $l_1$ and $l_2$. 
Subsequently, $l$ and $o_l$ will be removed and the new nodes,  and weights will be appended as follows: $\mathcal{I} = \mathcal{I}\cup \{i\}$, and $\mathcal{L} = \left(\mathcal{L}\cup \{l_1, l_2\}\right) \setminus \{l\}$, and $\mathcal{W} = \mathcal{W} \cup \{w_i\}$, and $\mathcal{O} = \left(\mathcal{O} \cup \{o_{l_1}, o_{l_2}\}\right) \setminus \{o_l\}$.
However, a sample may reach multiple leaf nodes due to the soft-routing mechanism, and thus the splits may differ from those of Hoeffding trees.
\begin{algorithm}[t]
   \caption{Attempt to split}
   \label{alg:attempt_to_split}
\begin{algorithmic}
    \STATE {\bfseries Input:} Leaf $l$, max\_depth, tie breaking threshold $\tau$
    \IF {samples seen so far are not from the same class \textbf{and} max\_depth not reached}
    \STATE Find split candidates by computing $\bar{G}(x_j)$ for each attribute $x_j$
    \STATE Select the top two $x_a$, $x_b$
    \STATE Compute Hoeffding bound $\epsilon$
    \IF{$\bigl((\bar{G}(x_a) - \bar{G}(x_b) =\Delta \bar{G}) > \epsilon$ \textbf{and} $x_a \neq x_{\emptyset}\bigr)$ \textbf{or} $\epsilon <$ $\tau$}
        \STATE Split $l$
    \ENDIF
\ENDIF
\end{algorithmic}
\end{algorithm}
After the tree has been traversed, gradient descent can be performed based on the computed gradients.

% Section: Transparency
\section{Transparency} \label{sec:transparency}
%In this section, we explain the transparency of a SoHoT and how we measure it compared to soft trees.
Post-hoc interpretations often fail to clearly explain how a model works. Therefore, we focus on transparency, aiming to elucidate the model's functioning rather than merely describing what else the model can tell the user \cite{Lipton2016}.
We examine the transparency of a SoHoT and how to measure it compared to soft trees at the level of the entire model (simulatability).

%\subsection{Transparency of SoHoT}
To make the model and the prediction transparent and explainable, we follow two principles: growth and inclusion of split tests. 
% 1. Growth
The growing tree structure enables a dynamic structure without prior knowledge of the incoming data stream. The user can see exactly when and how the structure changes based on the data stream and which split criterion is relevant, especially after a drift.
% 2. Transparent routing function
The second principle is based on the transparent routing function (see Eq.~\ref{eq:routing}). This takes into account the selected split criterion, which is easy for a user to read and understand.
Moreover, feature importance can be extracted from the model, as with each split, a feature is chosen that effectively reduces the impurity within the data, ultimately making a contribution to the overall prediction.
% Why is an ensemble still explainable?

%\subsection{Measure Transparency}\label{subsec:evaluatetradeoff}
We aim to compare the interpretability of the decision rules by the number of important features. 
The smaller the number of important features, the easier and shorter the explanation and therefore the more transparent the model.
We define a feature $i$ as important if its impact $\left|w_i x_i \right|$ on the decision rule exceeds an uniform distribution relative to $\langle w, x\rangle$.
A decision rule for soft Hoeffding trees and soft trees is based on $S(\langle w, x\rangle )$, where \mbox{$w=(w_1,w_2, ..., w_p)^T$}, \mbox{$x=(x_1, x_2, ..., x_p)^T$}.
If a summand in $w_1 x_1 + w_2 x_2 + \dots + w_p x_p$ is weighted more strongly (positively or negatively) by $w$, the corresponding feature has a greater influence on the decision rule $S(\langle w, x\rangle )$. Set $\sigma \coloneqq \left|w_1 x_1 \right| + \left|w_2 x_2 \right| + \dots + \left|w_p x_p \right|$.
The number of features that have more than average influence on the decision rule is defined by
\begin{align}\label{eq:tel_transparency}
    \sum_{i=1}^p \mathds{1} \biggl( \frac{\left|w_i x_i \right|}{\sigma} \geq \frac{1}{p} \biggr).
\end{align}

However, for SoHoT, the impact of  $S(\langle w, x\rangle )$ is weighted by $\alpha$.
The lower $\alpha$, the higher the impact just one feature has on the decision rule (see Eq.~\ref{eq:routing}).
To determine the number of features that impact the decision rule, the ratio of impact must be weighted by $\alpha$. 
The split criterion, which is weighted with $1-\alpha$, only considers one feature and therefore, the feature is ranked as an important feature depending on the $\alpha$ value.
Hence, the number of important features of a SoHoT is determined by
\begin{align}\label{eq:sohot_transparency_formular}
    \Biggl( \sum_{i=1}^p \mathds{1}\biggl(\alpha \cdot \frac{\left|w_i x_i \right|}{\sigma} \geq \frac{1}{p}\biggr) \Biggr) + \mathds{1} \Big( 1-\alpha \geq \frac{1}{p} \Big) .
\end{align}

% Evaluation: Trade-off
\begin{figure}[t!]
    \centering
    \begin{subfigure}[b]{0.4\textwidth}
        \scalebox{0.62}{
            \includegraphics[]{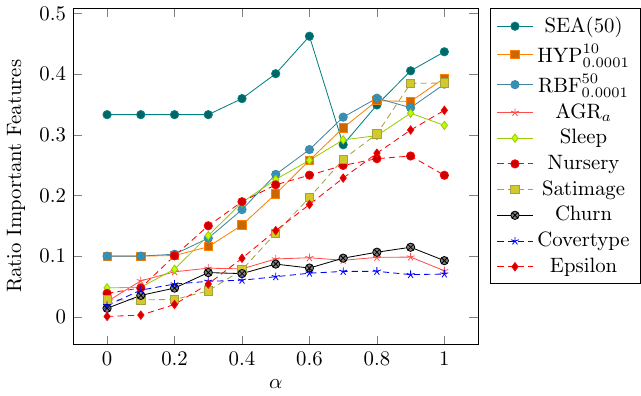}
        }
    \end{subfigure}
     \hspace{6em}
    \begin{subfigure}[b]{0.4\textwidth}
        \scalebox{0.62}{
            \includegraphics[]{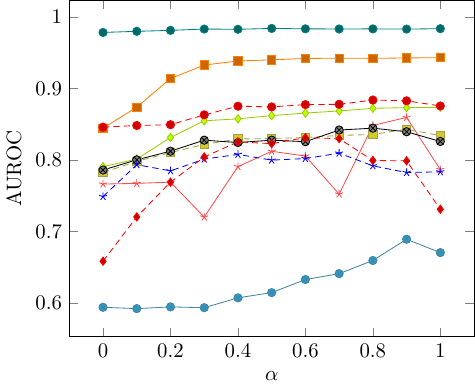}
        }
    \end{subfigure}
    \caption{Trade-off between transparency and performance for one SoHoT with test-then-train of a tree of depth 7. The results for the other data streams have very similar trends and are omitted due to space constraints.}
    \label{fig:tradeoff}
\end{figure}

As $\alpha$ regulates the impact of the univariate split criterion, varying $\alpha$ results in a trade-off between transparency and predictive performance. Small values for $\alpha$ yield more univariate splits, while values close to 1 yield more multivariate splits. Figure~\ref{fig:tradeoff} shows the average proportion of important features to the total number of features for the respective data stream and the AUROC for varying $\alpha$ on each data stream.
The transparency ratio peak for the SEA data stream occurs when the univariate split criterion weighting is deemed unimportant (Eq.~\ref{eq:sohot_transparency_formular}), due to the small number of features (3). The drop is also observed in other streams at different $\alpha$ values.

% Section: Evaluation
\section{Experiments}\label{sec:experiments}

We evaluate the performance of soft Hoeffding trees in terms of prediction and transparency by comparing SoHoTs to Hoeffding trees and soft trees.
First we evaluate how a SoHoT adapts to evolving data streams by analyzing the gradients and the behavior of growth.
The goal of the second experiment is to show that soft Hoeffding trees outperform Hoeffding trees at estimating class probabilities.
Finally, to clarify how SoHoT maintains transparency compared to soft trees, the third set of experiments compare the average ratio of important features on 20 data streams.
\begin{table}[t]
%\centering
\caption{Characteristics of the data streams, with e.g., abrupt (a) or gradual (g) drift, or injected perturbation (p). The streams are synthetic (s) generated or from real-world (r) sources.}
\label{table:Data-Description}
\scalebox{0.75}{
    \parbox{.4\linewidth}{
        \begin{tabular}{lcccccr}
        \toprule
        Data stream & Instances & Features & classes & Drift & Source\\
        \midrule
         SEA($50$)                          & $10^6$     & 3 & 2 & a & s\\
         SEA($5\cdot 10^5$)                 & $10^6$     & 3 & 2 & a & s \\
         $\text{HYP}^{10}_{0.0001}$         & $10^7$    & 10 & 2 & g & s \\
         $\text{HYP}^{10}_{0.001}$          & $10^7$    & 10 & 2 & g & s  \\
         $\text{RBF}^{50}_{0.0001}$         & $10^7$    & 10 & 5 & g & s \\
         $\text{RBF}^{50}_{0.001}$          & $10^7$    & 10 & 5 & g & s  \\
         $\text{AGR}_{a}$                   & $10^7$    & 9 & 2 & a & s \\
         $\text{AGR}_{p}$                   & $10^7$    & 9 & 2 & p & s \\
          \hline
         Poker                              & 1,025,010     & 10 & 10 & - & r & \\
         KDD99                              & 4,898,430     & 41 & 32 & - & r \\
        \bottomrule
        \end{tabular}
    }
}
\hspace{6em}
\scalebox{0.75}{
    \parbox{.4\linewidth}{
        \begin{tabular}{lcccccr}
        \toprule
        Data stream & Instances & Features & classes & Drift & Source\\
        \midrule
         Sleep                              & $10^6$     & 13 & 5 & a & r,s  \\
         Nursery                            & $10^6$     & 8 & 4 & a & r,s  \\
         Twonorm                            & $10^6$     & 20 & 2 & a & r,s  \\
         Ann-Thyroid                        & $10^6$     & 21 & 3 & a & r,s  \\
         Satimage                           & $10^6$     & 36 & 6 & a & r,s  \\
         Optdigits                          & $10^6$     & 64 & 10 & a & r,s  \\
         Texture                            & $10^6$     & 40 & 11  & a & r,s  \\
         Churn                              & $10^6$     & 20 & 2  & a & r,s  \\
          \hline
         Covertype                          & 581,010       & 54 & 7 & - & r  \\
         Epsilon                            & $10^5$       & 2,000 & 2 & - & r  \\
        \bottomrule
        \end{tabular}
    }
}
\end{table}
We employ 20 classification data streams (binary and multiclass), 8 of which purely synthetic, 4 are large real-world data streams (both are used in various literature \cite{2013PAW, 2022BifetHyperparameter}), and the remaining 8 are synthetic streams derived from real-world data streams (from the PMLB repository \cite{olson2017pmlb}) and generated as follows. We employ a Conditional Tabular Generative Adversarial Network (CTGAN) \cite{ctgan} to emulate a realistic distribution within a substantial data stream of $10^6$ samples and injected abrupt drifts by oversampling a randomly selected class. Ten drifts are induced, and each context contains samples in which around $75\%$ belong to the randomly selected class.

\subsection{Implementation Details}
We provide an open source Python implementation\footnote{\url{https://github.com/kramerlab/SoHoT}}. 
We assume a test-then-train setting, all measurements are averaged over 5 runs, on each run the data is randomly shuffled and the results are reported along with their standard errors.
SoHoTs were tested and trained using PyTorch \cite{pytorch2019}, utilizing the Adam optimizer \cite{KingBa15} and cross-entropy loss. As Hazimeh et al. \cite{hazimeh2020tree} discussed for TEL, we also precede SoHoT and ST by a batch normalization layer \cite{IoffeBN2015} and apply Eq. \ref{smoothstepfunc} as routing function for ST.
% Hyperparameter tuning
To obtain the benefits of the models, we employ an efficient per-instance training of a pool of models to enable hyperparameter tuning on streams \cite{CAND2022}. The model with the lowest estimated loss is chosen for every prediction and half of the models in the pool are selected for training. The hyperparameter selection is shown in Table~\ref{table:hyperparameter}.
\begin{table}[t]
\centering
\caption{Tuned hyperparameters and their range. Abbreviations: mc - majority class, nba - naive Bayes adaptive.}
\label{table:hyperparameter}
\begin{tabular}{llllr}
\toprule
SoHoT & HT & ST\\
\midrule
Max depth: $\{5, 6, 7\}$  &   Leaf prediction: $\{\text{mc}, \text{nba} \}$ &     Tree depth: $\{5, 6, 7\}$\\
$\gamma$: $\{1, 0.1\}$ \tablefootnote{To tune the data stream Hyperplane we use $\gamma$: $\{0.5, 0.1\}$ due to performance.}  &   $\delta$: $\{10^{-6}, 10^{-7}, 10^{-8}\}$ &   $\gamma$: $\{1, 0.1, 0.01\}$  \\
$\alpha$: $\{0.2, 0.3, 0.4\}$  & Grace period: $\{200, 400, 600\}$ & Learning rate: $\{10^{-2}, 10^{-3}\}$\\
\bottomrule
\end{tabular}
\end{table}
% Explain HT limit
To enable a fair comparison, we also compare SoHoT with a limited version of HT, $\text{HT}_{\text{limit}}$, which has a maximum of $(2^{7+1}-2) / 2 = 127$ internal nodes, matching the deepest SoHoT in the model pool.

\subsection{Evaluation}\label{subsec:evaluation}
% Structure:
%   1. Drift Adaption
%   2. Tables 
%       a. SoHoT vs. HT
%       b. SoHoT vs. ST
%   3. Transparency

% 1. Drift adaption
\begin{figure}[t!]
    \centering
    \begin{subfigure}[b]{0.4\textwidth}
        \scalebox{0.53}{
            \includegraphics{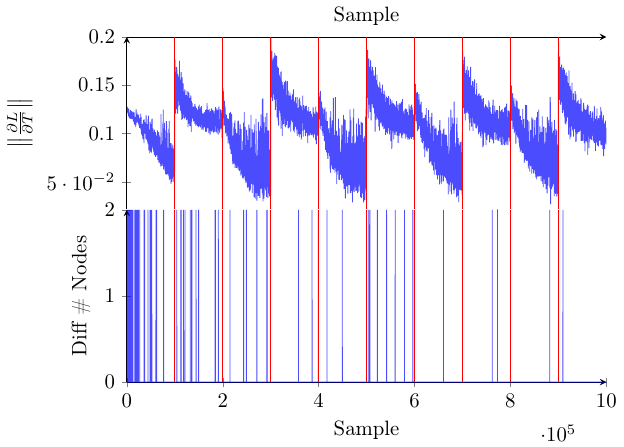}
        }
        \caption{Agrawal with abrupt drift.}
    \end{subfigure}
     \hspace{2.5em}
    \begin{subfigure}[b]{0.4\textwidth}
        \scalebox{0.53}{
            \includegraphics{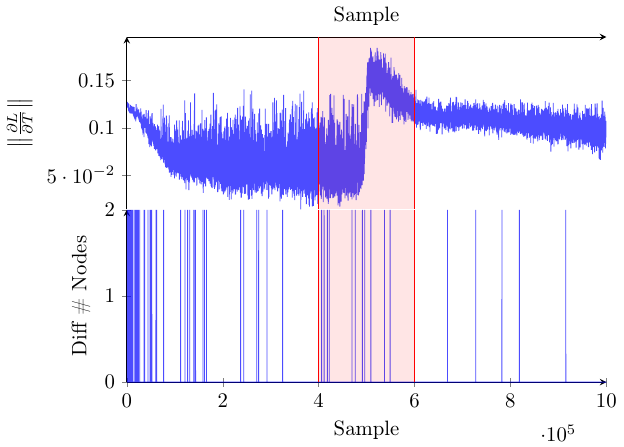}
        }
        \caption{Agrawal with gradual drift.}
    \end{subfigure}
    \caption{Drift adaptation for a single SoHoT on Agrawal stream with (a) abrupt and (b) gradual drift (indicated by red lines).
    The difference in number of nodes in a SoHoT and $\left\Vert\frac{\partial L}{\partial T}\right\Vert$, where $T$ is the output of the tree and $L$ the loss function, shows how a SoHoT adapts to different types of drifts over time.}
    \label{fig:drift_adaption}
\end{figure}
Soft Hoeffding trees adapt to drifting data streams by updating the weights in the tree by gradient descent and growing the tree structure by splitting leaf nodes. Figure~\ref{fig:drift_adaption} illustrates the extent to which a SoHoT adapts to a drifting data stream by visualizing $\left\Vert\frac{\partial L}{\partial T}\right\Vert$ and the number of added nodes over time.

%   2. Tables 
%       a. SoHoT vs. HT
\begin{table}[t]
\centering
\caption{SoHoT against HT with hyperparameter tuning on 20 data streams, averaged over five random repetitions and reported with standard error.
The $*$ indicates there is statistical significance based on a paired two-sided t-test at significance level of $0.05$.
}
\label{table:ht_multi}

\scalebox{0.73}{
    \begin{tabular}{l|ccc|cccr}
    \toprule
    Data stream                     & \multicolumn{3}{c}{Cross-entropy loss}  & \multicolumn{3}{c}{AUROC} \\
                         & SoHoT      & HT     & $\text{HT}_{\text{limit}}$  & SoHoT    & HT     & $\text{HT}_{\text{limit}}$ \\
    \midrule
    
SEA($50$) 	& \textbf{0.157}$^*$ $\pm$ 0.0014	& 3.841 $\pm$ 0.0326	& 3.810 $\pm$ 0.0304	& 0.984 $\pm$ 0.0002	& \textbf{0.994}$^*$ $\pm$ 0.0001	& \textbf{0.994} $\pm$ 0.0001\\
SEA($5\cdot 10^5$) 	& \textbf{0.157}$^*$ $\pm$ 0.0020	& 3.810 $\pm$ 0.0287	& 3.832 $\pm$ 0.0247	& 0.984 $\pm$ 0.0003	& \textbf{0.994}$^*$ $\pm$ 0.0002	& \textbf{0.994} $\pm$ 0.0001\\
$\text{HYP}^{10}_{0.0001}$ 	& \textbf{0.274}$^*$ $\pm$ 0.0004	& 0.348 $\pm$ 0.0112	& 0.374 $\pm$ 0.0173	& \textbf{0.947}$^*$ $\pm$ 0.0001	& 0.928 $\pm$ 0.0038	& 0.913 $\pm$ 0.0078\\
$\text{HYP}^{10}_{0.001}$ 	& \textbf{0.279}$^*$ $\pm$ 0.0018	& 0.397 $\pm$ 0.0156	& 0.433 $\pm$ 0.0185	& \textbf{0.946}$^*$ $\pm$ 0.0004	& 0.908 $\pm$ 0.0073	& 0.883 $\pm$ 0.0117\\
$\text{RBF}^{50}_{0.0001}$ 	& 1.474 $\pm$ 0.0094	& \textbf{1.473} $\pm$ 0.0125	& 1.488 $\pm$ 0.0110	& 0.628 $\pm$ 0.0025	& \textbf{0.660} $\pm$ 0.0106	& 0.629 $\pm$ 0.0099\\
$\text{RBF}^{50}_{0.001}$ 	& 1.525 $\pm$ 0.0096	& \textbf{1.523} $\pm$ 0.0100	& \textbf{1.523} $\pm$ 0.0098	& 0.532 $\pm$ 0.0040	& \textbf{0.546} $\pm$ 0.0046	& 0.545 $\pm$ 0.0048\\
$\text{AGR}_{a}$ 	& 0.139 $\pm$ 0.0266	& \textbf{0.081} $\pm$ 0.0199	& 0.084 $\pm$ 0.0199	& 0.978 $\pm$ 0.0085	& 0.993 $\pm$ 0.0014	& \textbf{0.994} $\pm$ 0.0008\\
$\text{AGR}_{p}$ 	& 0.435 $\pm$ 0.0003	& \textbf{0.389}$^*$ $\pm$ 0.0002	& \textbf{0.389} $\pm$ 0.0001	& 0.855 $\pm$ 0.0002	& \textbf{0.883}$^*$ $\pm$ 0.0001	& \textbf{0.883} $\pm$ 0.0001\\
\hline
Sleep 	& \textbf{0.975} $\pm$ 0.0121	& 0.986 $\pm$ 0.0158	& 0.977 $\pm$ 0.0188	& \textbf{0.852} $\pm$ 0.0033	& 0.851 $\pm$ 0.0037	& 0.852 $\pm$ 0.0041\\
Nursery 	& \textbf{0.806} $\pm$ 0.0177	& 0.820 $\pm$ 0.0228	& 0.820 $\pm$ 0.0224	& \textbf{0.868} $\pm$ 0.0059	& 0.867 $\pm$ 0.0020	& 0.867 $\pm$ 0.0020\\
Twonorm 	& 0.165 $\pm$ 0.0092	& 0.135 $\pm$ 0.0072	& \textbf{0.126} $\pm$ 0.0056	& 0.986 $\pm$ 0.0015	& 0.987 $\pm$ 0.0009	& \textbf{0.988} $\pm$ 0.0008\\
Ann-Thyroid 	& \textbf{0.598} $\pm$ 0.0210	& 0.637 $\pm$ 0.0240	& 0.647 $\pm$ 0.0252	& \textbf{0.844} $\pm$ 0.0098	& 0.830 $\pm$ 0.0119	& 0.825 $\pm$ 0.0123\\
Satimage 	& \textbf{1.214} $\pm$ 0.0238	& 1.327 $\pm$ 0.0365	& 1.328 $\pm$ 0.0357	& \textbf{0.803}$^*$ $\pm$ 0.0117	& 0.772 $\pm$ 0.0071	& 0.771 $\pm$ 0.0074\\
Optdigits 	& \textbf{1.768} $\pm$ 0.0186	& 1.780 $\pm$ 0.0189	& 1.779 $\pm$ 0.0187	& \textbf{0.711} $\pm$ 0.0028	& 0.710 $\pm$ 0.0042	& 0.711 $\pm$ 0.0040\\
Texture 	& \textbf{1.516} $\pm$ 0.0412	& 1.598 $\pm$ 0.0578	& 1.592 $\pm$ 0.0556	& \textbf{0.770} $\pm$ 0.0069	& 0.760 $\pm$ 0.0095	& 0.762 $\pm$ 0.0082\\
Churn 	& \textbf{0.448} $\pm$ 0.0347	& 0.494 $\pm$ 0.0354	& 0.490 $\pm$ 0.0351	& \textbf{0.805} $\pm$ 0.0223	& 0.776 $\pm$ 0.0225	& 0.780 $\pm$ 0.0235\\
\hline
Poker 	& 1.079 $\pm$ 0.0008	& \textbf{0.978}$^*$ $\pm$ 0.0002	& \textbf{0.978} $\pm$ 0.0001	& 0.501 $\pm$ 0.0072	& 0.597 $\pm$ 0.0066	& \textbf{0.601} $\pm$ 0.0043\\
Covertype 	& 1.025 $\pm$ 0.0046	& \textbf{0.760}$^*$ $\pm$ 0.0035	& \textbf{0.760} $\pm$ 0.0018	& 0.790 $\pm$ 0.0019	& \textbf{0.901}$^*$ $\pm$ 0.0013	& 0.900 $\pm$ 0.0010\\
Kdd99 	& 0.090 $\pm$ 0.0036	& \textbf{0.026}$^*$ $\pm$ 0.0001	& \textbf{0.026} $\pm$ 0.0001	& 0.877 $\pm$ 0.0048	& \textbf{0.904}$^*$ $\pm$ 0.0024	& \textbf{0.904} $\pm$ 0.0028\\
Epsilon 	& \textbf{0.580}$^*$ $\pm$ 0.0066	& 0.668 $\pm$ 0.0011	& 0.668 $\pm$ 0.0011	& \textbf{0.838}$^*$ $\pm$ 0.0076	& 0.665 $\pm$ 0.0012	& 0.665 $\pm$ 0.0012\\

    \hline
    \# Wins  & 12    & 7  & 6  & 10    & 7  & 7 \\
    
    \bottomrule
    \end{tabular}
}
\end{table}
To evaluate the ability to predict class probabilities, we analyze the cross-entropy loss.
Table~\ref{table:ht_multi} compares the average cross-entropy loss and AUROC of \mbox{SoHoT}, HT, and $\text{HT}_{\text{limit}}$ across 20 data streams.
SoHoT outperforms HT and $\text{HT}_{\text{limit}}$ on 12 data streams in terms of cross-entropy loss.
Even for AUROC, SoHoT outperforms both Hoeffding tree variations on 10 data streams.
SoHoT offers better adaptability on the large data stream Hyperplane containing multiple gradual drifts.
For the data streams in the middle section of Table~\ref{table:ht_multi} (Sleep to Churn), we can observe SoHoT is particularly effective for streams with imbalanced class sampling, especially when the predominant class abruptly changes over time.
Overall, the results show that SoHoT outperforms HT and $\text{HT}_{\text{limit}}$ in 12 out of the 20 cases, and is outperformed in 6-7 cases, at the task of estimating class probabilities.

%       b. SoHoT vs. ST
%Average absolute deviation AUROC:  0.04606556267552109
%Average absolute deviation CE Loss:  0.1368941573589146
%Average relative error AUROC: 5.2911 %
%Average relative error CE Loss: 53.1124 %

\begin{table}[t]
\centering
\caption{Performance comparison of SoHoT against ST with hyperparameter tuning on 20 data streams, averaged over five random repetitions and reported with standard error.
The $*$ indicates there is statistical significance based on a paired two-sided t-test at significance level of $0.05$.
}
\label{table:tel_multi}

\scalebox{0.8}{

    \begin{tabular}{l|cc|ccr}
    \toprule
    Data stream                     & \multicolumn{2}{c}{Cross-entropy loss}  & \multicolumn{2}{c}{AUROC} \\
                         & SoHoT      & ST  & SoHoT    & ST \\
    \midrule
    
SEA($50$) 	& \textbf{0.157} $\pm$ 0.0014	& 0.182 $\pm$ 0.1025	& \textbf{0.984} $\pm$ 0.0002	& 0.914 $\pm$ 0.0760\\
SEA($5\cdot 10^5$) 	& 0.157 $\pm$ 0.0020	& \textbf{0.067}$^*$ $\pm$ 0.0002	& 0.984 $\pm$ 0.0003	& \textbf{0.999}$^*$ $\pm$ 0.0000\\
$\text{HYP}^{10}_{0.0001}$ 	& 0.274 $\pm$ 0.0004	& \textbf{0.216}$^*$ $\pm$ 0.0002	& 0.947 $\pm$ 0.0001	& \textbf{0.959}$^*$ $\pm$ 0.0000\\
$\text{HYP}^{10}_{0.001}$ 	& 0.279 $\pm$ 0.0018	& \textbf{0.216}$^*$ $\pm$ 0.0002	& 0.946 $\pm$ 0.0004	& \textbf{0.959}$^*$ $\pm$ 0.0000\\
$\text{RBF}^{50}_{0.0001}$ 	& 1.474 $\pm$ 0.0094	& \textbf{0.915}$^*$ $\pm$ 0.0102	& 0.628 $\pm$ 0.0025	& \textbf{0.884}$^*$ $\pm$ 0.0021\\
$\text{RBF}^{50}_{0.001}$ 	& 1.525 $\pm$ 0.0096	& \textbf{1.515} $\pm$ 0.0097	& 0.532 $\pm$ 0.0040	& \textbf{0.557}$^*$ $\pm$ 0.0060\\
$\text{AGR}_{a}$ 	& 0.139 $\pm$ 0.0266	& \textbf{0.053}$^*$ $\pm$ 0.0086	& 0.978 $\pm$ 0.0085	& \textbf{0.998} $\pm$ 0.0008\\
$\text{AGR}_{p}$ 	& 0.435 $\pm$ 0.0003	& \textbf{0.395}$^*$ $\pm$ 0.0001	& 0.855 $\pm$ 0.0002	& \textbf{0.882}$^*$ $\pm$ 0.0001\\
\hline
Sleep 	& 0.975 $\pm$ 0.0121	& \textbf{0.733}$^*$ $\pm$ 0.0127	& 0.852 $\pm$ 0.0033	& \textbf{0.915}$^*$ $\pm$ 0.0024\\
Nursery 	& 0.806 $\pm$ 0.0177	& \textbf{0.713}$^*$ $\pm$ 0.0138	& 0.868 $\pm$ 0.0059	& \textbf{0.894}$^*$ $\pm$ 0.0043\\
Twonorm 	& 0.165 $\pm$ 0.0092	& \textbf{0.069}$^*$ $\pm$ 0.0049	& 0.986 $\pm$ 0.0015	& \textbf{0.996}$^*$ $\pm$ 0.0005\\
Ann-Thyroid 	& 0.598 $\pm$ 0.0210	& \textbf{0.523}$^*$ $\pm$ 0.0272	& 0.844 $\pm$ 0.0098	& \textbf{0.860}$^*$ $\pm$ 0.0096\\
Satimage 	& 1.214 $\pm$ 0.0238	& \textbf{1.058}$^*$ $\pm$ 0.0268	& 0.803 $\pm$ 0.0117	& \textbf{0.847}$^*$ $\pm$ 0.0095\\
Optdigits 	& 1.768 $\pm$ 0.0186	& \textbf{1.674}$^*$ $\pm$ 0.0186	& 0.711 $\pm$ 0.0028	& \textbf{0.726}$^*$ $\pm$ 0.0042\\
Texture 	& 1.516 $\pm$ 0.0412	& \textbf{1.120}$^*$ $\pm$ 0.0639	& 0.770 $\pm$ 0.0069	& \textbf{0.850}$^*$ $\pm$ 0.0080\\
Churn 	& 0.448 $\pm$ 0.0347	& \textbf{0.428}$^*$ $\pm$ 0.0369	& 0.805 $\pm$ 0.0223	& \textbf{0.814}$^*$ $\pm$ 0.0236\\
\hline
Poker 	& 1.079 $\pm$ 0.0008	& \textbf{0.986}$^*$ $\pm$ 0.0001	& \textbf{0.501} $\pm$ 0.0072	& 0.491 $\pm$ 0.0095\\
Covertype 	& 1.025 $\pm$ 0.0046	& \textbf{0.703}$^*$ $\pm$ 0.0031	& 0.790 $\pm$ 0.0019	& \textbf{0.915}$^*$ $\pm$ 0.0010\\
Kdd99 	& 0.090 $\pm$ 0.0036	& \textbf{0.024}$^*$ $\pm$ 0.0005	& 0.877 $\pm$ 0.0048	& \textbf{0.909}$^*$ $\pm$ 0.0044\\
Epsilon 	& 0.580 $\pm$ 0.0066	& \textbf{0.426}$^*$ $\pm$ 0.0012	& 0.838 $\pm$ 0.0076	& \textbf{0.889}$^*$ $\pm$ 0.0016\\

    \hline
    \# Wins  & 1    & 19  & 2  & 18  \\
    
    \bottomrule
    \end{tabular}
}
\end{table}
Next, we examine our transparent and incremental trees in comparison to soft trees.
The mean cross-entropy loss and AUROC along with the standard error are shown in Table~\ref{table:tel_multi}. 
%Statistical significances are omitted, as ST is significantly better for all but 2 (3) data streams in cross-entropy loss (AUROC).
As expected, ST has a lower cross-entropy loss and a higher AUROC than SoHoT. 
This is due to the transparent gating function and the changing tree complexity. Regarding the latter point, newly added weights after a split must first be adjusted to the data, which is not initially required for soft trees.
The cross-entropy loss for SoHoT is $29.21 \%$ higher on the median compared to ST.
The results regarding AUROC exhibit similar trends to those observed for cross-entropy loss.
The AUROC for ST is $2.88 \%$ better on the median compared to SoHoT's performance. 
Note that SoHoT's performance should approach the one of ST with $\alpha=1$, once the tree is fully grown.

%   3. Transparency      
\begin{figure}
    \centering
        \scalebox{0.9}{
            \includegraphics[]{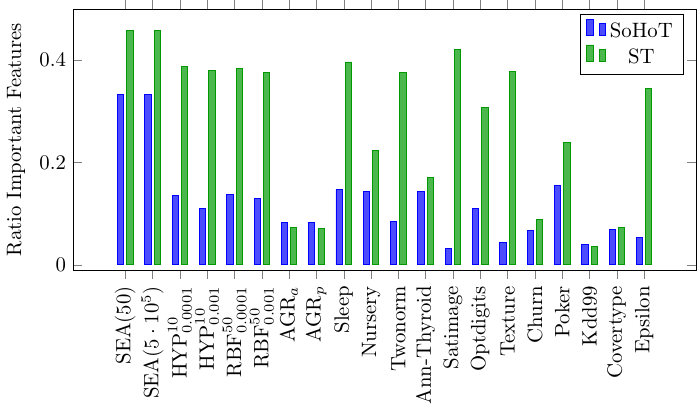}
        }

    \caption{The average proportion of important features per decision rule in a single SoHoT with $\alpha=0.3$ and a soft tree (ST) for $n=10^5$ samples of each data stream.}
    \label{fig:measuretransparency}
\end{figure}

Finally, we explore the advantage SoHoTs provide in terms of transparency.
In Section~\ref{sec:transparency}, we proposed a metric to evaluate the transparency of soft Hoeffding trees and soft trees (ST).
We compare the transparency of our gating function and the smooth-step function using 20 data streams. 
Therefore, we analyze the average number of important features per decision rule for one single soft Hoeffding tree and one single soft tree.
For SoHoT, $\alpha$ is set to $0.3$.
Figure~\ref{fig:measuretransparency} visualizes the average proportion of important features to the total number of features for the respective data stream.
It can be observed that on average the number of important features for a decision rule is higher for soft trees than for SoHoTs.
This suggests that the coding length of an explanation for SoHoT is shorter and therefore easier than for soft trees.
To regulate transparency, $\alpha$ can be employed by assigning a higher weight to a single feature from the split test. However, this adjustment may lead to a potential loss of performance, as illustrated in Figure~\ref{fig:tradeoff}.

% Section: Conclusion
\section{Conclusion}\label{sec:conclusion}
We introduced soft Hoeffding trees as a transparent and differentiable model, which utilizes a new transparent routing function.
Our tree can handle large data streams and is able to adapt to drifting streams by growing new subtrees and by updating node weights.
Our experiments indicate that SoHoT is often better at estimating class probabilities in comparison to Hoeffding trees with unlimited and limited depth.
Soft Hoeffding trees trade-off transparency and performance with an adjustable parameter. Soft trees perform better than SoHoT, but the transparent structure and explainable routing mechanism of SoHoTs provide them with a distinct advantage over soft trees.
We have shown that SoHoTs provide a shorter explanation of feature importance than soft trees, which is beneficial for the transparency of SoHoTs.
A drawback of SoHoTs in comparison to Hoeffding trees and soft trees is the amount of hyperparameters.
We like to highlight that our algorithm can also be applied to regression tasks. To do so, the split test computation simply needs to be adjusted, such as by minimizing variance in the target space \cite{breiman1984classification}.
In this case, the output dimension becomes $k=1$.
% Discussion on Hoeffding bound
Rutkowski et al. criticized the application of Hoeffding's inequality in mining data streams and presented the use of McDiarmid's bound instead \cite{Rutkowski2013}. In this work, the bound was applied heuristically, as in many other machine learning papers using stochastic bounds, and the McDiarmid's bound could similarly be employed as an alternative. 
% Future work
Future investigations will focus to design the splits more in response to the changes in the data stream, i.e. updating the split tests based on the samples that reach the internal node such as in Extremely Fast Decision Trees \cite{ExtremelyFastDT2018}.
Quantifying transparency is challenging, so the next steps would be to conduct a user study \cite{doshi2017towards}.

The focus of this paper was on the soft Hoeffding tree as an individual model, compared to other individual models. As a next step, we are going to evaluate ensembles of SoHoTs in comparison to ensembles of HTs and STs.

% Note use of \abovespace and \belowspace to get reasonable spacing
% above and below tabular lines.

\begin{credits}
\subsubsection{\ackname} The research in this paper was supported by the ``TOPML: Trading Off Non-Functional Properties of Machine Learning'' project funded by Carl Zeiss Foundation, grant number P2021-02-014.
\end{credits}

%
% ---- Bibliography ----
%
% BibTeX users should specify bibliography style 'splncs04'.
% References will then be sorted and formatted in the correct style.
%
\bibliographystyle{splncs04}
\bibliography{sohot}

\end{document}